\def\year{2020}\relax
\documentclass[letterpaper]{article} 
\usepackage{aaai20}  
\usepackage{times}  
\usepackage{helvet} 
\usepackage{courier}  
\usepackage{amsmath,bm}
\newtheorem{theorem}{Theorem}
\newtheorem{definition}{Definition}
\newtheorem{proposition}{Proposition}

\newcommand{\KL}{D_{\mathrm{KL}}}

\newcommand{\citet}[1]{\citeauthor{#1} \shortcite{#1}}

\def\vx{{\bm{x}}}
\def\vy{{\bm{y}}}

\usepackage{algorithm}
\usepackage{algorithmic}
\usepackage[hyphens]{url}  
\usepackage{graphicx} 
\usepackage{graphicx}  
\usepackage{xcolor}
\usepackage{subcaption}
\usepackage{booktabs} 
\usepackage{amsfonts}

\title{Infomax Neural Joint Source-Channel Coding via Adversarial Bit Flip}

\author{
\Large \textbf{
Yuxuan Song$^1$,
Minkai Xu$^1$,
Lantao Yu$^2$,
Hao Zhou$^3$,
Shuo Shao$^1$,
Yong Yu$^1$
}\\
\textsuperscript{\rm 1} Shanghai Jiao Tong University,
\textsuperscript{\rm 2} Stanford University
\textsuperscript{\rm 3} Bytedance AI lab
\\
\{songyuxuan,mkxu,yyu\}@apex.sjtu.edu.cn,
lantaoyu@cs.stanford.edu,\\
zhouhao.nlp@bytedance.com, shuoshao@sjtu.edu.cn
}
\begin{document}

\maketitle

\begin{abstract}

Although Shannon theory states that it is asymptotically optimal to separate the source and channel coding as two independent processes, in many practical communication scenarios this decomposition is limited by the finite bit-length and computational power for decoding.
Recently, neural joint source-channel coding (NECST)  \cite{choi2018necst} is proposed to sidestep this problem. While it leverages the advancements of amortized inference and deep learning \cite{kingma2013auto,grover2018uncertainty} to improve the encoding and decoding process, it still cannot always achieve compelling results in terms of  compression and error correction performance due to the limited robustness of its learned coding networks.
In this paper, motivated by the inherent connections between neural joint source-channel coding and discrete representation learning, we propose a novel regularization method called Infomax Adversarial-Bit-Flip (IABF) to improve the stability and robustness of the neural joint source-channel coding scheme. More specifically, on the encoder side, we propose to explicitly maximize the mutual information between the codeword and data; while on the decoder side, the amortized reconstruction is regularized within an adversarial framework. Extensive experiments conducted on various real-world datasets evidence that our IABF can achieve state-of-the-art performances on both compression and error correction benchmarks and outperform the baselines by a significant margin. 

\end{abstract}

\section{Introduction}
Shannon source-channel separation theorem~\cite{shannon1948mathematical} is one of the most fundamental theorems in information theory. It states that when the source and channel coding procedures are conducted separately there is an asymptotically negligible loss in the point-to-point communication system, which indicates the system can be simplified as a series of two subsystems without any interference to each other.
However, although the separation approach is optimal in many scenarios \cite{tian2013optimality}, it can be proved to be sub-optimal when the source generates finite-length data blocks \cite{kostina2013lossy}, which is common in practical communication scenarios. 
Furthermore, in this case, maximum likelihood decoding was proven to be an NP-hard problem \cite{berlekamp1978inherent}, and relaxation approaches of this problem \cite{feldman2005using,vontobel2007low} also suffer from high distortion rate issues.

To address the above-mentioned challenges, the machine learning community has cast the encoding process of joint source-channel coding as a binary representation learning problem. 
Recently, inspired by variational mutual information maximization, \cite{choi2018necst} proposed Neural Error Correcting and Source Trimming (NECST), which is a deep learning framework for joint source-channel coding. To be more specific, the data is encoded with a network into a bit-string representation without the need of manually designing the coding scheme. Additionally, based on amortized optimization, the framework provides an extremely fast decoder after training.

However, while NECST leverages the advancements of amortized inference and deep learning to improve the compression and reconstruction process,  it still cannot always achieve compelling results when concerning about compression and error correction performance, which is mainly due to the limited informativeness and robustness of learned coding networks. In this paper, we propose a novel approach to solve the previous problems in NECST via mutual information maximization and amortized regularization~\cite{shu2018amortized}. On the encoder side, we employ the idea of mutual information maximization for the better informativeness, which is predominant in representation learning \cite{hu2017learning,zhao2017infovae,rolfe2016discrete}. However, we note that the previous mutual information maximization methods in neural discrete representation learning are intractable when the dimension (\emph{i.e.} length of bit-string) is high. To address this issue, we further propose a novel loss function which can be estimated and optimized with deep neural networks more effectively. On the decoder side, we theoretically demonstrate that the objective in \cite{choi2018necst} is essentially regularizing the amortized decoder and the strength of regularization is determined by the noise level of the channel. The amortized optimization tends to encourage the decoder to capture more global structures during the decoding, and it is reasonable to be suitably regularized. Based on this intuition, we further proposed ``Adversarial Bits Flip'', a new regularization mechanism for discrete input neural network to improve the robustness of coding scheme.

More specifically, our contributions can be summarized as follows: 
\begin{itemize}
	\item We propose a novel mutual information maximization framework, which is scalable for high-dimensional discrete representation learning;
	\item Theoretical analysis of the learning paradigm in neural joint source-channel coding is conducted. Based on the theoretical understanding, we propose a novel virtual adversarial regularization method to improve the robustness of the discrete-input decoder networks;
	\item We conduct extensive experiments on various benchmark datasets and different tasks. Empirical evidence demonstrates the effectiveness of our methods on both reducing the distortion with finite bit-length and learning useful representation for downstream tasks.
\end{itemize}

\section{Related Work}
Generative models have been regarded as the backbones of both lossless and lossy compression methods. Recently, there are a surge of studies on applying deep generative models to lossy compression.  \cite{balle2018variational} proposed to substitute the fixed prior in Variational Auto-encoder (VAE) with a learnable scale hyper-prior such that the structure information can be better presented for effective compression.  \cite{theis2017lossy} utilized autoencoding framework to obtain the optimal number of bits for compressing images. Likelihood-free methods based on adversarial learning are also proposed to learn neural codes for better compression \cite{santurkar2018generative}. 

The robustness of the learned coding scheme is another important topic under the communication setting. Here we discuss two recent progress in the aspect of the involvement of channel noise.  \cite{grover2018uncertainty} proposed Uncertainty Autoencoder (UAE) to learn the compressed representation of the original inputs, and \cite{choi2018necst} can be seen as a discrete version of UAE, \emph{i.e.}, both the codewords and the noise model is no longer continuous. Both  \cite{grover2018uncertainty} and \cite{choi2018necst} maximize the mutual information within a variational framework, while in our work, we tend to enhance the mutual information between the codewords and the input data without introducing any variational approximation. 
In the literature of information-theoretic approach for representation learning,  \cite{chen2016infogan,hjelm2018learning,van2017neural,zhao2018information} also proposed to utilize information maximization to improve representation learning. However, in their settings, the discrete latent noise is not involved. Specifically,  \cite{hu2017learning} introduced a component for regularizing the encoder function with Virtual Adversarial Training (VAT) \cite{miyato2015distributional}. Although both inspired by VAT, our method instead regularizes the decoder function. The underlying motivations are also different: \cite{hu2017learning} tend to impose intended invariance on discrete representations, while our method aims to enhance the robustness of coding scheme by stimulating and improving the worst-case performance for some channel noise level.



\section{Background and Notations}
\subsection{Joint Source and Channel Coding}
To begin with, we firstly formulate the problem of communicating data across a noisy channel. Following the notations in  \cite{choi2018necst}, we denote the input space as $\mathcal{X} \subseteq \mathbb{R}^n$, and the source distribution on the input space as $p_\text{data}(\vx)$. The communication system encodes a block of data $\vx_{1:n}$ \emph{i.i.d.}  drawn from $p_\text{data}(\vx)$  into codewords. Specifically, for each data instance $\vx$, the corresponding codeword $\vy$ is a binary code of length $m$, \emph{i.e.}, $\vy_{1:m} \in \mathcal{Y}= \{0,1\}^{m}$. The codeword will then be transmitted through a noisy channel, which results in a \emph{corrupted} codeword $\hat{\vy}_{1:m} \in \mathcal{Y}$.
After receiving the noisy codes, the decoder will produce a reconstructed version $\hat{\vx} \in \mathcal{X}$.  The ultimate goal is to minimize the overall distortion, \emph{i.e.}, minimizing the $\ell_p$ norm $\|x-\hat{x}\|_p$(typically $p=1$ or $2$).

Source encoder tries to compress source message into a  bit-string with as less number of bits as possible. While channel encoder tends to re-introduces redundancies for error correction after the transmission through noisy channel. Shannon proved that the above scheme is optimal for infinitely long messages in the separation theorem~\cite{shannon1948mathematical}. While in practice, when the bit-length is finite, a joint source-channel coding algorithm can have a better performance than the separated source-channel coding~\cite{pilc1967coding}.

\subsection{Autoencoder and Variational Information Maximization}
Autoencoder \cite{ballard1987modular} consists of a pair of parameterized functions, an encoder ($f_\theta$) and a decoder ($f_\phi$). The encoder maps sample $\vx$ from the $n$-dimensional data space $\mathcal{X}$ to  a codeword $\vy$ in the $m$-dimensional latent space $\mathcal{Y}$, and the decoder defines a function from the latent space to the data space. Typically, an autoencoder seeks to minimize the $l_2$ reconstruction error over a dataset $\mathcal{D}$:
\begin{equation}
\label{AE_loss}
\min _{f_\theta , f_\phi} \sum_{x \in \mathcal{D}}\|x-f_\phi(f_\theta(x))\|_{2}^{2}.
\end{equation}
Usually both $f_\phi$ and $f_\theta$ are parameterized with neural networks.
When probabilistic encoder and decoder are utilized, the encoding part and the decoding part essentially imply corresponding conditional distributions, \emph{i.e.} $p_\theta(\vy|\vx)$ and $q_\phi(\vx|\vy)$. 
And the corresponding joint distribution $p_\theta(\vx,\vy)$ between the two random variables is also defined by the factorization $p_\theta(\vx,\vy)=p_\theta(\vy|\vx)p_\text{data}(\vx)$.
When the objective is to obtain informative measurements to reconstruct the original signal effectively, it is reasonable to maximize the mutual information $I_\theta(X,Y)$ between these two random variables $X$ and $Y$:
\begin{align}
\max _{\theta} I_{\theta}(X, Y)&=\int p_{\theta}(\vx, \vy) \log \frac{p_{\theta}(\vx, \vy)}{p_{\text {data }}(\vx) p_{\theta}(\vy)} \mathrm{d} x \mathrm{d} y \nonumber\\ 
&=H(X)-H_{\theta}(X | Y)
\end{align}
Here $H$ stands for differential entropy. Estimating and optimizing the mutual information between high-dimensional random variables is intractable. However, the mutual information can actually be lower bounded by introducing a variational approximation of the posterior $p_\theta(\vx|\vy)$. With $q_\phi(\vx|\vy)$ representing the variational distribution, the lower bound can be written as:
\begin{equation}
H(X)+\mathbb{E}_{p_{\theta}(\vx, \vy)}[\log q_\phi(\vx | \vy)] \leq I_{\theta}(X, Y)
\end{equation}
The bound is tight when the variational distribution $q_\phi(\vx|\vy)$ matches the true posterior $p_\theta(\vx|\vy)$. Since $H(X)$ is the entropy of data distribution, it can be treated as a constant during the optimization of $\theta$ and $\phi$. The final objective of the stochastic optimization can be concluded as \cite{grover2018uncertainty}:
\begin{equation}
\label{VBO}
\max _{\theta, \phi} \mathbb{E}_{p_{\theta}(\vx, \vy)}\left[\log q_\phi(\vx | \vy)\right]
\end{equation}

\subsection{Neural Error Correcting and Source Trimming Codes}
Let $X$,$Y$,$\hat{Y}$,$\hat{X}$ the random variables for the inputs, codewords, noisy codewords after channel corruption, and the reconstructed data by decoder. 
\cite{choi2018necst} modeled a coding process with the following graphical model $X \rightarrow Y \rightarrow \hat{Y} \rightarrow \hat{X}$:
\begin{equation}
\begin{array}{l}{p(\vx, \vy, \hat{\vy}, \hat{\vx})=} \\ {p_{\text {data}}(\vx) p_\theta(\vy| \vx) p_{\text {channel}}(\hat{\vy} | \vy ; \epsilon) q_\phi(\hat{\vx} | \hat{\vy})}\end{array}
\end{equation}
Here $p_{\text {channel}}(\hat{\boldsymbol{y}} | \boldsymbol{y} ; \epsilon)$ denotes the probabilistic model of the noisy channel, \emph{i.e.}, flipping each bit with probability $\epsilon$.
Inspired by recent advancements of latent-variable generative models,  \cite{choi2018necst} proposed a variational information maximization method based on Eq.~\ref{VBO} to improve the information theoretic dependency between the input $X$ and the noisy codeword $\hat{Y}$ in joint source-channel coding.

\begin{figure}[!t]
\centering     
\includegraphics[width=0.60\columnwidth]{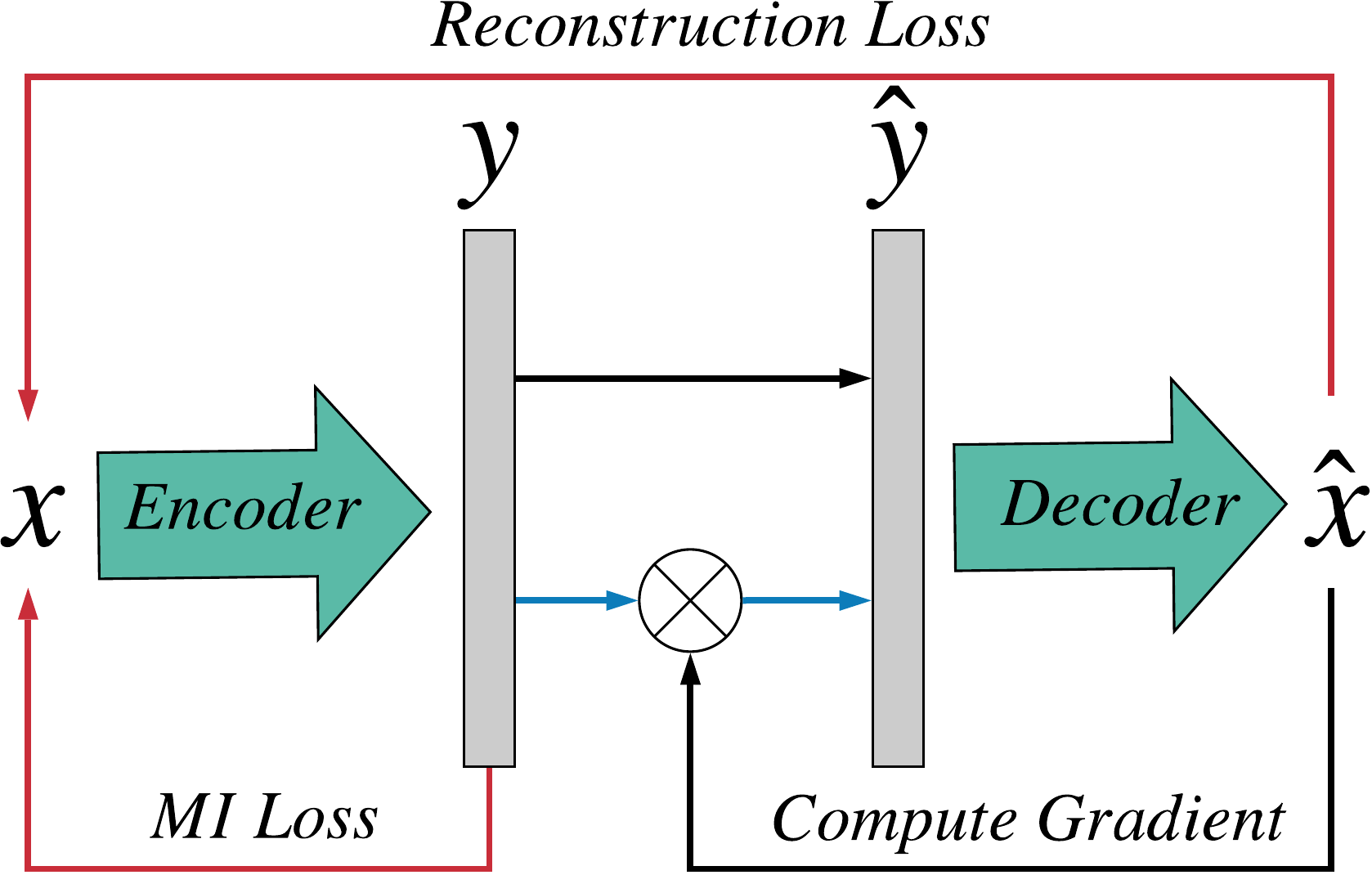}
\caption{Overview of our proposed IABF algorithm. Green: encoder and decoder networks. $\vx$ denotes the input data and $\hat{\vx}$ denotes the reconstructed data from coding. $\vy$ denotes the codeword before the noisy channel and $\hat{\vy}$ denotes the noisy codeword. Training procedure: First (black arrows), we encode the input data $\vx$ to $\vy$, then we directly decode the clean code $\vy$ and compute the reconstruction loss. According to the reconstruction loss, we compute the gradients of different bits. Secondly (blue arrows), we flip the bits according to the gradient norm and obtain the noisy code $\hat{\vy}$. Finally (red arrow), we compute the reconstruction loss between $\vx$ and $\hat{\vx}$ and the mutual information loss between $\vx$ and $\vy$, and then update the parameters of encoder and decoder networks according to the two loss components.}
\label{fig:model_overview}
\end{figure}

\section{Methodology}

\subsection{Motivation}
Optimizing the variational bound as proposed in \cite{choi2018necst} can approximately maximize the corresponding mutual information between the data and noisy codewords. 
However, it should be noticed that previous variational approximation based method is mainly limited by the capacity of the parameterized variational distribution family \cite{kingma2013auto} and high-variance gradient estimation.
In this paper, we take an alternative perspective: the maximization of $I(X,\hat{Y})$ can be decomposed into maximizing $I(X,Y)$ and minimizing the information loss during noisy channel corruption. Based on the above decomposition, we propose a novel method to sidestep the limitations of variational approximation and further improve the error correction ability. To be more specific, $I(X,Y)$ is maximized without involving a parameterized variational distribution and the robustness of a coding scheme is strengthened in an adversarial fashion for reducing information loss.

In the following, we first introduce our method on how to directly impose mutual information maximization without involving variational distribution in joint source-channel coding. Then we conduct a theoretical study on the learning paradigm of the NECST where we demonstrate that injecting latent noise essentially smooths the amortized decoder function. 
Based on the analysis, we propose a regularization method "Adversarial Bit Flip", which can effectively regularize the decoder by virtually attacking the vulnerable bits during training, which further improve the robustness of the learned coding scheme. 

\subsection{Information Maximization with Discrete Representation}
\label{infomax}
The sample space of $M$-bit codeword can be denoted as $\mathcal{Y}=\mathcal{Y}_1 \times \cdots \times \mathcal{Y_{M}}$, where $\mathcal{Y}_m\equiv\left\{0,1\right\}$ and $Y_m$ stands for the random variable for the $m$-th bit in codewords.
We seek to learn a probabilistic encoder $p_\theta(\vy_1,\cdots,\vy_M |\vx)$ which implies the optimal coding strategy. Following previous works \cite{kingma2013auto,grover2018uncertainty}, we also leverage the mean-field assumption when modeling the conditional distribution with neural networks, \emph{i.e.}
\begin{equation}
\label{mf}
p_{\theta}\left(\vy_{1}, \ldots, \vy_{M} | \vx\right)=\prod_{d=1}^{M} p_{\theta}\left(\vy_{d} | \vx\right)
\end{equation}
 
The mutual information $I(Y_1,\cdots,Y_M;X)$ between the codewords and inputs is intractable for large bits number $M$, as the summation over an exponential number of terms is involved. 
Following  \cite{brown2009new}, we derive the mutual information between a set of codewords and inputs as the following sum of Interaction Information terms:
\begin{equation}
I\left(Y_{1 : M} ; X\right)=\sum_{T \subseteq S, |T| \geq 1} I(T \cup Y)
\end{equation}
where $S\equiv \left\{Y_1,\cdots,Y_M\right\}$. With similar approximation by truncating all the higher order terms with $|T|>2$ \cite{hu2017learning,brown2009new,erin2015deep}, the new approximated target can be illustrated as:
\begin{align}
\label{expansion}
\sum_{d=1}^{M} I\left(Y_{d} ; X\right)+\sum_{1 \leq d \neq d^{\prime} \leq M} I\left(\left\{Y_{d}, Y_{d^{\prime}}, X\right\}\right)    
\end{align}

Note that under the mean-field assumption in Eq.~\ref{mf}, the pairwise interaction information can be derived as:
\begin{align}
\label{eq:9}
I\left(\left\{Y_{d}, Y_{d^{\prime}}, X\right\}\right) &\equiv I\left(Y_{d} ; Y_{d^{\prime}} | X\right)-I\left(Y_{d} ; Y_{d^{\prime}}\right) \nonumber\\
&=-I\left(Y_{d} ; Y_{d^{\prime}}\right)
\end{align} 
Substituting Eq.~\ref{expansion} into Eq.~\ref{eq:9}, we get the approximated mutual information maximization term:
\begin{equation}
\label{eq:M}
I\left(Y_{1 : M} ; X\right) = \sum_{d=1}^{M} I\left(X ; Y_{d}\right)-\sum_{1 \leq d \neq d^{\prime} \leq M} I\left(Y_{d} ; Y_{d^{\prime}}\right)
\end{equation}
The final objective consists of two terms. The first term $\sum_{d=1}^{M} I\left(X ; Y_{d}\right) = \sum_{d=1}^{M}(H(Y_d)-H(Y_d| X)) $ corresponds to maximizing the the summation of mutual information between data and each bit of codewords, which can be easily derived under the situation of binary representation:
\begin{align}
\label{hyx}
&H(Y_d) \equiv h\left(p_{\theta}(\vy_d)\right)=H\left(\frac{1}{N} \sum_{i=1}^{N} p_{\theta}(\vy_d| \vx^{(i)})\right)\nonumber\\
&H(Y_d| X) \equiv \frac{1}{N} \sum_{i=1}^{N} H\left(p_{\theta}\left(\vy_d | \vx^{(i)}\right)\right)
\end{align}
where $H(\cdot)$ denotes the entropy of a discrete distribution and $N$ stands for the number of samples.
However, the second term in Eq.~\ref{eq:M} corresponds to the orthogonality constraint which dedicates to removing the redundancy between different dimension of codewords. 
To impose the orthogonality constraint on different dimensions of codewords, instead of directly optimizing the intractable target in Eq.~\ref{eq:M}, we introduce Total Correlation as an independence measure of multi-dimensional random variable:
\begin{equation}
TC (Y_{1 : M})= \KL\left(p_\theta(\vy_{1: M}) \| \prod_{j=1}^{M} p_\theta \left(\vy_{j}\right)\right)
\end{equation}
Samples from marginal distribution $p_\theta$ can be easily obtained through the following ancestral sampling procedure: 
$\vx \sim p(\vx), \vy_{1:M} \sim p_\theta(\vy_{1:M}|\vx)$. And sampling for distribution $\prod_{j=1}^{M} p_\theta \left(\vy_{j}\right)$ can be implemented through randomly permuting across a batch of samples from $p_\theta(\vy_{1:M})$ for each dimension. To optimize $TC (Y_{1 : M})$, we leverage the density ratio estimation trick as illustrated in  \cite{kim2018disentangling}. We mix up the samples from $p_\theta(\vy_{1: M})$  and  $\prod_{j=1}^{M} p_\theta \left(\vy_{j}\right)$, and train a MLP classifier to output the probability $C_\psi$ of a codeword coming from $p_\theta(\vy_{1: M})$ with the following objective:
\begin{equation}
\begin{aligned} 
     &\mathbb{E}_{\vy_{1:M} \sim p_\theta(\vy_{1:M})} [\log C_\psi(\vy_{1:M})] +\\
     &\mathbb{E}_{\vy_{1:M} \sim \prod_{j=1}^{M} p_\theta \left(\vy_{j}\right) } [\log(1-C_\psi(\vy_{1:M}))]
\end{aligned}
\end{equation}

And the total correlation can be approximated as \cite{goodfellow2014distinguishability}:
\begin{align}
\label{tc}
TC (\vy_{1:M})&= \mathbb{E}_{p_\theta(\vy_{1:M})}\left[\log \frac{p_\theta(\vy_{1:M})}{\prod_{j=1}^{M} p_\theta \left(\vy_{j}\right)}\right]\nonumber\\
&\approx \mathbb{E}_{p_\theta(\vy_{1:M})}\left[\log \frac{C_\psi(\vy_{1:M})}{1-C_\psi(\vy_{1:M})}\right] = \hat{TC} (\vy_{1:M})
\end{align}
\subsection{Amortized Decoder Regularization with Adversarial Bit Flip}
\label{bitflip}
Given variational family $\mathcal{Q}$ and the joint distribution $p_\theta(\vx,\vy)$ implicitly defined by the encoder function $p_\theta(\vy|\vx)$ and $p_\text{data}(\vx)$, the variational mutual information maximization can be formalized as:
\begin{align}
    \max _{\theta} \mathbb{E}_{p_\text{data}(\vx) p_\theta(\boldsymbol{y}|\vx)}[&H(p_\theta(\cdot|\boldsymbol{y}))-\nonumber\\
    &\min _{q \in \mathcal{Q}} \KL\left(p_{\theta}(\vx|\boldsymbol{y})\|q(\vx))\right)]
\end{align}
where $\KL(\cdot||\cdot)$ denotes the Kullback-Leibler divergence. It can be shown that variational bound approximates the original objective best when we use the best decoding distribution $q^{*}_{\boldsymbol{y}}(x)$ for each $\boldsymbol{y} \sim p_\theta(\boldsymbol{y})$. The amortized optimization \cite{choi2018necst} turns the individual optimization procedure for finding the optimal decoding distribution for each codeword $\boldsymbol{y}$ into a single regression problem by using a recognition model $f_\phi: \mathcal{Y} \rightarrow \mathcal{Q}$ to predict $q^{*}_{\boldsymbol{y}}(x)$. And the function $f_\phi$ can be concisely represented as the conditional distribution $q_\phi(x|\boldsymbol{y})$ which results in the form of objective in Eq.~\ref{VBO}. The amortization is appealing in neural joint source-channel coding as the amortized decoder can very efficiently map the transmitted code into its best reconstruction at test time. However, amortization with the neural network as the variational function family is over-expressive in many cases and prone to overfitting \cite{shu2018amortized}, while the objective of joint source-channel coding is to find the coding scheme which can generalize to the unseen test data and achieve the desired compression performance. Hence regularizing the capacity of amortization family can be seen as the way to refine the decoder module to match the expected desiderata.

\subsubsection{NECST as an Amortized Decoder Regularization}
In NECST, the final objective is to maximize the variational bound of mutual information between the corrupted codewords and the input data:
\begin{equation}
\max _{\theta, \phi} \mathbb{E}_{\vx \sim p_{\text {data}}(\vx)} \mathbb{E}_{\vy \sim p_\text{noisy}(\vy | \vx ; \epsilon, \theta)}\left[\log q_\phi(\vx | \vy)\right]
\end{equation}
where
\begin{equation}
p_\text{noisy}(\vy | \vx;\epsilon, \theta)= \sum_{\hat{\vy} \in \hat{\mathcal{Y}}} p_{\theta}(\hat{\vy} | \vx) p_{\text{channel}}(\vy | \hat{\vy} ; \epsilon)
\end{equation}
is defined  according to a different noise channel. 
Following the typical setting of joint source-channel coding, we briefly introduce two widely used discrete channel models: (1) the binary erasure channel (BEC); and (2) the binary symmetric channel (BSC). In BEC, each bit may be i.i.d. erased into an unrecognized symbol with some probability $\epsilon$, and the uncorrupted bits will be transmitted faithfully. In BSC, each bit will be flipped independently with probability $\epsilon$, (\emph{e.g.} $0\rightarrow 1$). It is noted that BSC is widely recognized as a more difficult communication channel than BEC \cite{richardson2008modern}. Hence in the following sections and experiments, we mainly focus on the scenario of BSC. And BEC can be easily adapted from the discussion on BSC. 
In the BSC channel coding, the whole encoding distribution can be formulated as following \cite{choi2018necst}:
\begin{equation}
\begin{aligned}
&p_\text{noisy}(\vy | \vx ; \theta, \epsilon)\\ =&\prod_{i=1}^{m}\left(\sigma\left(f_{\theta}\left(\vx\right)_{i}\right)-2 \sigma\left(f_{\theta}\left(\vx\right)_{i}\right) \epsilon+\epsilon\right)^{\vy_{i}} \cdot \\ &(1-\sigma\left(f_{\theta}\left(\vx\right)_{i}\right)+2 \sigma\left(f_{\theta}\left(\vx\right)_{i}\right) \epsilon-\epsilon)^{\left(1-\vy_{i}\right)} \end{aligned}
\end{equation}
where each bit $\vy_i$ is modeled as an independent Bernoulli distribution, the parameter of the Bernoulli distribution is modeled through a neural network $f_\theta$, and $\sigma$ denotes the Sigmoid function. 
While from the perspective of the decoder module, the injected noise defines a new learning objective for amortized learning:
\begin{align}
\label{NECST}
\max _{\theta}\min_{\phi}~&I_{\theta}(X, Y)-\\
&\mathbb{E}_{p_\text{data}(\vx)p_\text{noisy}(\vy | \vx; \theta, \epsilon)}[\KL(p_{\theta}(\vx | \vy)\| f_\phi(\vy))]\nonumber
\end{align}

We then show that the optimal amortized decoder with the channel noise is the following kernel function. And the noise level $\epsilon$ actually determine the smoothness of the optimal decoder.
\begin{definition}\label{def:kernel}
The univariate kernel of two discrete random variable is defined as following:
\begin{equation}
K_{1}^{(U)}\left(y_{i}, y_{j}\right)=\left\{\begin{array}{c}{P_{Y}\left(y_{i}\right) \text { if } y_{i}=y_{j}} \\ {0 \quad\quad~~~ \text { if } y_{i} \neq y_{j}}\end{array}\right.
\end{equation}
where $P_Y$ is the probability mass function of a discrete random variable $Y$.
And the corresponding multi-variate kernel is:
\begin{equation}
K\left(\boldsymbol{y}^{(i)}, \boldsymbol{y}^{(j)}\right)=\frac{1}{m} \sum_{k=1}^{m} k_{1}^{(U)}\left(\boldsymbol{y}^{(i)}_k, \boldsymbol{y}^{(j)}_k\right),
\end{equation}
\end{definition}
In our case the perturbed distribution $P_Y = \sigma\left(f_{\theta}\left(\vx\right)_{i}\right)-2 \sigma\left(f_{\theta}\left(\vx\right)_{i}\right) \epsilon+\epsilon$ for $Y=1$ and $P_Y = 1-\sigma\left(f_{\theta}\left(\vx\right)_{i}\right)+2 \sigma\left(f_{\theta}\left(\vx\right)_{i}\right) \epsilon-\epsilon$ for $Y=0$. We refer to the corresponding kernel with noise level $\epsilon$ as $K_\epsilon$.
\begin{theorem}
\label{kernel}
The optimal amortized decoder given a fixed encoder in Eq.~\ref{NECST} is a kernel regression model, which can be illustrated as:
{\small
\begin{equation}
f_{\epsilon,\phi}^{*}(\vy)=\underset{f_\phi \in \mathcal{F}(q)}{\arg \min } \sum_{i=1}^{n} w_{\epsilon}\left(\vy, \vy^{(i)}\right) \cdot \KL\left(p_{\theta}\left(\vx | \vy^{(i)}\right)\|f_\phi(\vy)\right)\nonumber
\end{equation}
}
where $w_{\epsilon}\left(\vy, \vy^{(i)}\right)=\frac{K_{\epsilon}\left(\vy, \vy^{(i)}\right)}{\sum_{j} K_{\epsilon}\left(\vy, \vy^{(j)}\right)}$; $K_\epsilon$ denote the kernel function defined in Definition \ref{def:kernel} with $\epsilon$ as the noise level; $n$ denotes the number of training samples.
\end{theorem}

Theorem.~\ref{def:kernel} states that the optimal amortized decoder is related to the noise level $\epsilon$. And the optimal $f_{\epsilon,\phi}^{*}(\vy)$ corresponds to the decoding procedure which minimizes the weighted Kullback-Leibler(KL) divergence from $f_{\epsilon,\phi}^{*}(\vy)$ to each $p_\theta(\vx|\vy)$. The weighting function $w_{\epsilon}\left(\vy, \vy^{(i)}\right)$ depends on the Bernoulli parameter of decoder output $\sigma\left(f_{\theta}\left(\vx\right)_{i}\right)$ and the noise level $\epsilon$.
It can be found that the noise level $\epsilon$ forces the decoder to get similar reconstruction outputs  with similar codewords under a predefined similarity measure. When the channel noise reach the highest level ,\emph{i.e.} $\epsilon = 0.5$, the weighted function will then turn into constant $w_{\epsilon}\left(\vy, \vy^{(i)}\right)= \frac{1}{n}$.  Intuitively, the noise level is related to the smoothness of the decoder function: how much a single decoding procedure will be influenced by the other decoding procedures. 

Formally, we have the following proposition:

\begin{proposition}
\label{p1}
With a regularized amortization objective as:
\begin{align}
    R&(\phi;\epsilon) =\nonumber\\
    &\min _{f_\phi \in \mathcal{F}(q)} \mathbb{E}_{p (\vx)p_{\text{noisy}}(\vy | \vx ; \theta, \epsilon)} [\KL(p_{\theta}(\vx | \vy) \| f_{\phi}(\vy))]
\end{align}
The following condition is satisfied if the input is closed under the permutation: if $f_\phi(\vy) \in \mathcal{F}(q)$ then the noised version $f_\phi(\hat{\vy};\epsilon) \in \mathcal{F}(q)$, where $f_\phi(\hat{y};\epsilon)$ denotes perturb the input $\hat{\vy}$ with noise channel $p_{\text{channel}}(\hat{\vy}|\vy;\epsilon)$.  Then it is satisfied that when $0 \leq \epsilon_1< \epsilon_2 < 0.5, R(\phi;\epsilon_1) \leq  R(\phi;\epsilon_2)$ for all $\phi \in \Phi$. 
\end{proposition}

Theorem.~\ref{kernel} and Proposition.~\ref{p1} show that with a larger noise level, the decoder is further regularized  and forced to be smoother. And the objective with smaller $\epsilon$ in Eq.~\ref{NECST} is necessarily bounded by the objective with larger $\epsilon$. So the regularized objectives are  valid lower bounds of the original variational bound of mutual information in Eq.~\ref{VBO}, where the $\epsilon$ can be seen as 0.  
\begin{algorithm}[t]
  \caption{Infomax Adversarial Bits Flip}
  \label{alg:mairl}
  \begin{algorithmic}[1]
  \STATE {\bfseries Input:} Dataset($\mathcal{X}$) to be compressed. Channel noise level $\epsilon$.

  \STATE Initialize the parameters of encoder $p_\theta(\vy_1,\dots,\vy_M|\vx)$, the parameter of decoder $q_\phi(\vx|\vy_1,\dots,\vy_M)$ and a Classifier $C_\psi$.

  \REPEAT
  \STATE Sample a batch of samples from Dataset: $\vx\sim p_\text{data}(\vx)$ 

  \STATE Sample the corresponding latent codes $\boldsymbol{y}\sim p_\theta(\boldsymbol{y}|\vx)$.
  \STATE Permute $\boldsymbol{y}$ across the batch  of  samples for each dimension to get $\boldsymbol{y}^{*}$.
  \STATE Update $\psi$ according to the classification loss:
  \STATE { \quad\quad$\log C_\psi(\boldsymbol{y}) + \log(1-C_\psi(\boldsymbol{y}^{*}))$}
  \STATE Update $\theta$, $\phi$ according to the objective in Eq.~\ref{obj}:
  \STATE  \quad\quad$\mathcal{L}_{rec}(\phi, \theta ; \vx, \epsilon) + \lambda \mathcal{L}_{info}(\theta ; \vx)$
  \UNTIL{Convergence}
  \STATE {\bfseries Output:} Learned joint source-channel coding scheme $p_\theta(\boldsymbol{y}|\vx)$ and amortized decoder $q_\phi(\vx|\boldsymbol{y})$
  \end{algorithmic}
\end{algorithm}

\subsubsection{Adversarial Bit Flip}
Above theoretical analysis helps us build connections between  neural joint source-channel coding and regularized amortized optimization.
Hence there is strong motivation to design effective regularization strategy for the amortized decoder and improve the stability and generalization performance of the coding scheme. In the continuous setting, the intended regularization of neural networks can be imposed through local perturbations \cite{miyato2015distributional,bachman2014learning}, where the neural network is encouraged to be invariant towards the perturbation. More specifically, in Virtual Adversarial Training (VAT) \cite{miyato2015distributional}, the perturbation is selected as the adversarial direction based on first-order gradient.
However, in our scenario, we tend to inject noise on the binary codewords which makes the gradient-based perturbation strategy no longer applicable. 
Inspired by VAT, we introduce the following ``adversarial bit flip'' procedure. 
Given a $M$-bit codeword $\vy \sim p_\theta(\vy|\vx)$, we first calculate the gradient with respect to the reconstruct loss $\mathcal{L} = \log q_\phi(\vx|\vy)$:
\begin{equation}
\nabla_{\vy} \mathcal{L}=\left[\frac{\partial \mathcal{L}}{\partial \vy_{1}}, \ldots, \frac{\partial \mathcal{L}}{\partial \vy_{M}}\right]
\end{equation}
The naive perturbation is to get the corrupted bits $\hat{\vy}$ by:
\begin{equation}
\label{abf}
\hat{\boldsymbol{y}}=\boldsymbol{y}+\operatorname{sign}\left(\nabla_{\boldsymbol{y}} \mathcal{L}\right)
\end{equation}
where the $\operatorname{sign}\left(\nabla_{\boldsymbol{y}} \mathcal{L}\right) \in \{-1,+1\}^M$. Since the codewords is constrained to be binary, perturbation following Eq.~\ref{abf} may output numbers not in $\{0,1\}$. Note that the bit flip should only happen in two situations, when the original bits is $0$ and the corresponding gradient is positive or the original bit is $1$ and the corresponding gradient is negative. Then we refine the perturbation procedure as following:
\begin{equation}
\label{maskcode}
\boldsymbol{m}=\boldsymbol{y} \oplus\left(\operatorname{sign}\left(\nabla_{\boldsymbol{y}} \mathcal{L}\right) / 2+0.5\right)
\end{equation}
\begin{equation}
\label{code_flip}
\hat{\boldsymbol{y}}=\boldsymbol{y} \oplus \boldsymbol{m}
\end{equation}
where $\oplus$ denotes the XOR operation. 
The key motivation of above procedure lies in that we leverage the gradient information as guidance to find the most \emph{vulnerable} bits and virtually attack these bits, which will implicitly force the information uniformly-distributed along different dimensions. From another perspective, the Adversarial Bit Flip can be seen as simulating the worst possible case given noise level fixed, which can provide more informative signal to improve the robustness. The whole training procedure is summarized in Algorithm~\ref{alg:mairl}.

\section{Experiments}
In this section, we firstly introduce the implementation and optimization in details. Then following the evaluation framework in  \cite{choi2018necst}, results on several benchmark datasets: BinaryMNIST, MNIST \cite{lecun1998mnist}, Omniglot \cite{lake2015human} and CIFAR10 \cite{krizhevsky2009learning} are provided to demonstrate the effectiveness of our methods on compression and error correction.\footnote{The codebase for this work can be found at \url{https://github.com/MinkaiXu/neural-coding-IABF}.}.  
\subsection{Objective and Optimization Procedure}
With the definition in Eq.~\ref{hyx} and Eq.~\ref{tc}, the mutual information maximization term for increasing $I(X,Y)$ is:
\begin{equation}
    \sum_{d=1}^{M}(H(Y_d)- H(Y_d|X)) - \hat{TC}(Y_{1 : M}) \equiv \mathcal{L}_{info}(\theta ; \vx)
\end{equation}
And the adversarial bits flip term can be illustrated as:
\begin{equation}
\sum_{\vx \in \mathcal{D}} \mathbb{E}_{\vy \sim p_{adv}(\vy | \vx ; \epsilon, \theta)}\left[\log q_\phi(\vx | \vy)\right] \equiv \mathcal{L}_{rec}(\phi, \theta ; \vx, \epsilon)
\end{equation}
Combining the above two components, we derive the final objective of our Infomax Adversarial Bits Flip(IABF) as following:
\begin{equation}
\label{obj}
    \max_{\theta,\phi} \mathcal{L}_{rec}(\phi, \theta ; \vx, \epsilon) + \lambda \mathcal{L}_{info}(\theta ; \vx)
\end{equation}
\begin{table}[!b]
\begin{tabular}{c|cccc}
\toprule
\textbf{Binary MNIST} & 0.1 & 0.2 & 0.3 & 0.4\\ 
\hline
\specialrule{0em}{0pt}{3pt}
100-bit NECST & 0.116 & 0.150 & 0.193 & 0.249\\
100-bit ABF & 0.114 & 0.149 & 0.191 & 0.244\\
100-bit IABF & \textbf{0.113} & \textbf{0.147} & \textbf{0.189} & \textbf{0.241} \\
\midrule
\textbf{Omniglot}  & 0.1 & 0.2 & 0.3 & 0.4\\
\hline
\specialrule{0em}{0pt}{3pt}
200-bit NECST & 24.693 & 31.538 & 39.657 & 47.888\\
200-bit ABF & 24.192 & 31.430 & 39.039 & 47.827\\
200-bit IABF & \textbf{24.132} & \textbf{31.373} & \textbf{38.936} & \textbf{47.608} \\
\midrule
\textbf{CIFAR10}  & 0.1 & 0.2 & 0.3 & 0.4\\ 
\hline
\specialrule{0em}{0pt}{3pt}
500-bit NECST & 63.238 & 74.402 & 89.619 & 117.803\\
500-bit ABF & 58.992 & 71.338 & \textbf{83.192} & 116.874\\
500-bit IABF & \textbf{55.351} & \textbf{70.525} & 83.193 & \textbf{116.281}\\
\midrule
\textbf{MNIST} & 0.1 & 0.2 & 0.3 & 0.4\\ 
\hline
\specialrule{0em}{0pt}{3pt}
100-bit NECST & 14.439 & 22.556 & 34.377 & 48.291\\
100-bit ABF & 13.369 & \textbf{22.013} & 33.319 & 48.041\\
100-bit IABF & \textbf{13.251} & 22.026 & \textbf{33.176} & \textbf{46.555} \\
\bottomrule
\end{tabular}
\caption{$\mathcal{L}^2$ squared reconstruction error loss (per image) of NECST vs. IABF. The error is calculated on test set.}
\label{compression_results}
\end{table}

where $\lambda$ is the only hyperparameter and selected from a small candidate set $\{0.1,0.01,0.001\}$ during the experiments.
The perturbed distribution implied by adversarial bit flip $p_{adv}(\vy | \vx ; \epsilon, \theta)$ is approximated with a continuous relaxation of Eq.~\ref{code_flip}:
\begin{equation}
\begin{aligned} 
&p_{adv}(\vy | \vx ; \theta,\epsilon)\\ =&\prod_{i=1}^{m}\left(\sigma\left(f_{\theta}\left(\vx\right)_{i}\right)-2 \sigma\left(f_{\theta}\left(\vx\right)_{i}\right) \epsilon_i^*+\epsilon_i^\cdot\right)^{\vy_{i}}\cdot \\
&\left(1-\sigma\left(f_{\theta}\left(\vx\right)_{i}\right)+2 \sigma\left(f_{\theta}\left(\vx\right)_{i}\right) \epsilon_i^*-\epsilon_i^*\right)^{\left(1-\vy_{i}\right)} \end{aligned}
\end{equation}
Instead of directly applying the same $\epsilon$ level noise on all the dimensions, 
here the noise level on $i$-dimension $\epsilon_i \propto \|\frac{\partial \mathcal{L}}{\partial \vy_{i}}\|$ satisfying that $\sum_{d=1}^{M}\epsilon_i = M\epsilon , 0\leq\epsilon_i \leq1$. And $\epsilon_i^*$ indicates the modified noise level with with the mask code $\boldsymbol{m}$ as shown in Eq.~\ref{maskcode}, \emph{i.e} $\epsilon_i^* = \epsilon_i \cdot m_i$.

Another practical challenge lies in the optimization procedure of $\mathcal{L}_{rec}(\phi, \theta ; \vx, \epsilon)$, the gradient of the encoder parameters $\theta$ is non-trivial to estimate. Hence we adapted  VIMCO \cite{mnih2016variational,choi2018necst}, a multi-sample variational lower bound objective for obtaining low-variance gradients.
Following \cite{choi2018necst}, we also use the multi-sample objective with $K=5$:
\begin{equation}
\begin{array}{l}{\mathcal{L}^{K}_{rec}(\phi, \theta ; \vx, \epsilon)=} \\ { \sum_{\vx \in \mathcal{D}} \mathbb{E}_{\vy^{1 : K} \sim p_{adv}(\vy | \vx ; \epsilon, \theta)}\left[\log \frac{1}{K} \sum_{i=1}^{K} q_{\phi}(\vx | \vy^{(i)})\right]}\end{array}
\end{equation}

\begin{figure}[!t]
\centering     
\begin{subfigure}{0.49\linewidth}
	\centering
	\includegraphics[width=0.99\columnwidth]{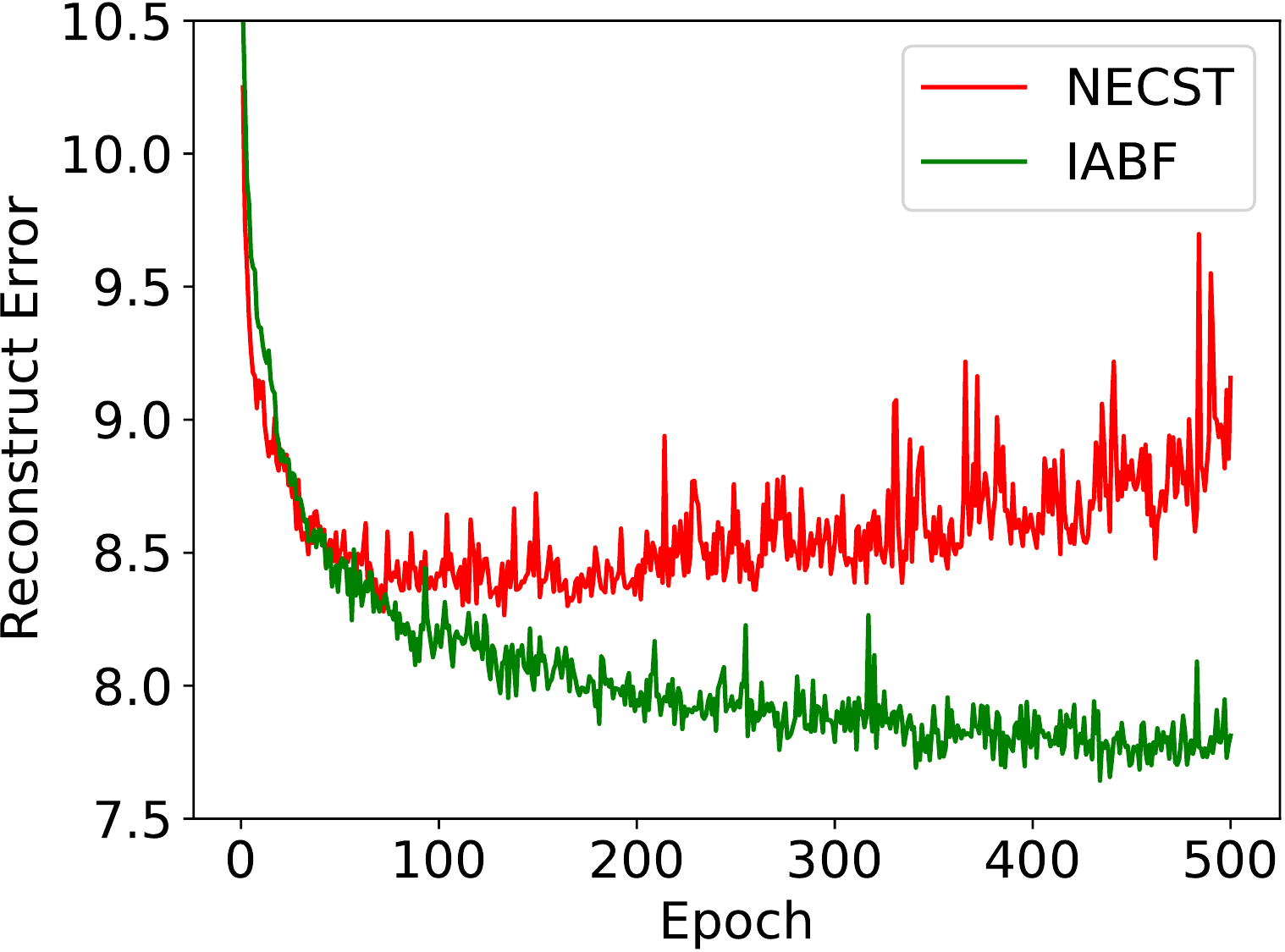}
	\caption{noise level: 0.1}
	\label{fig:noise_1}
\end{subfigure}
\begin{subfigure}{0.49\linewidth}
	\centering
	\includegraphics[width=0.99\columnwidth]{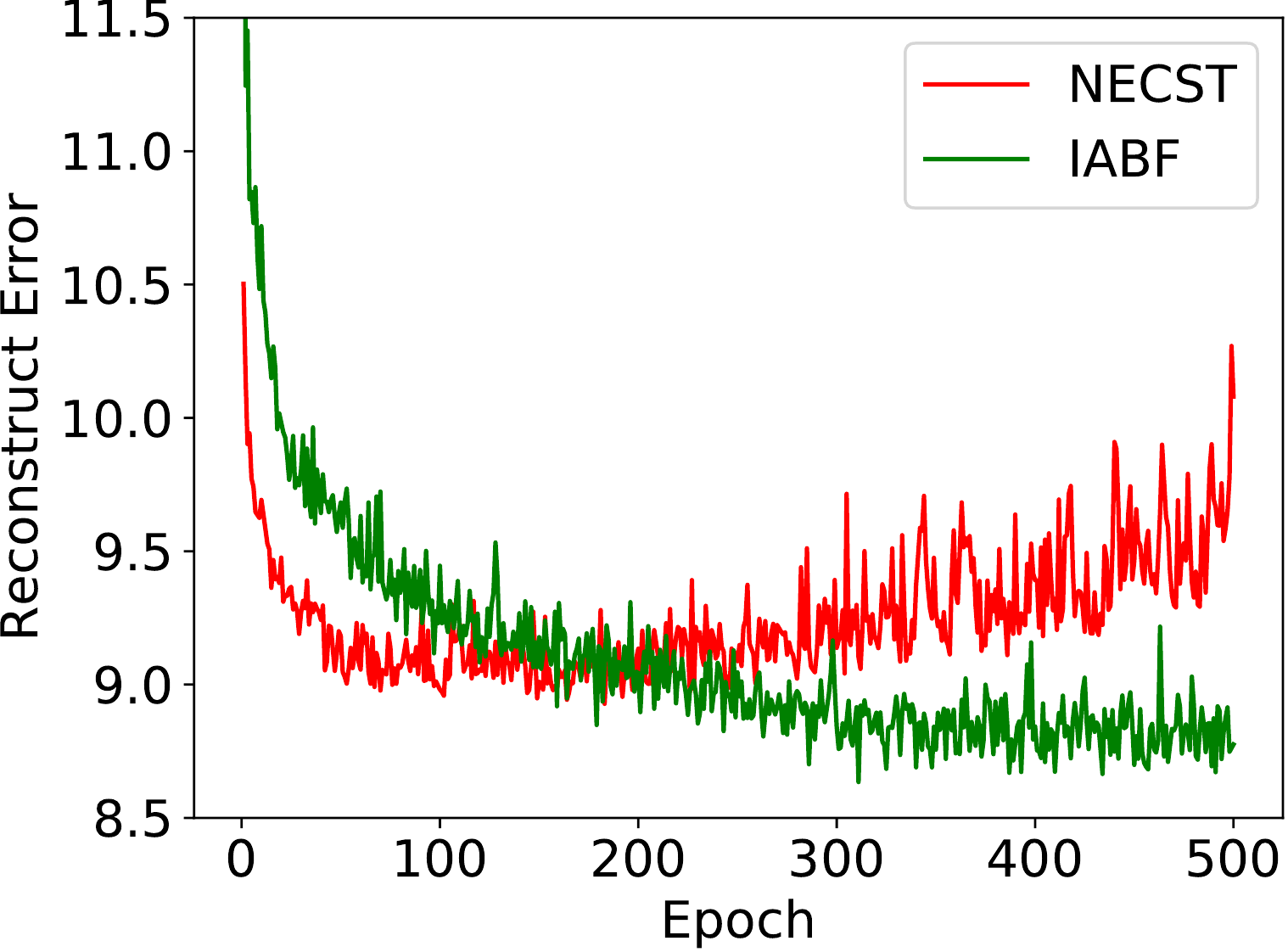}
	\caption{noise level: 0.2}
	\label{fig:noise_2}
\end{subfigure}
\begin{subfigure}{0.49\linewidth}
	\centering
	\includegraphics[width=0.99\columnwidth]{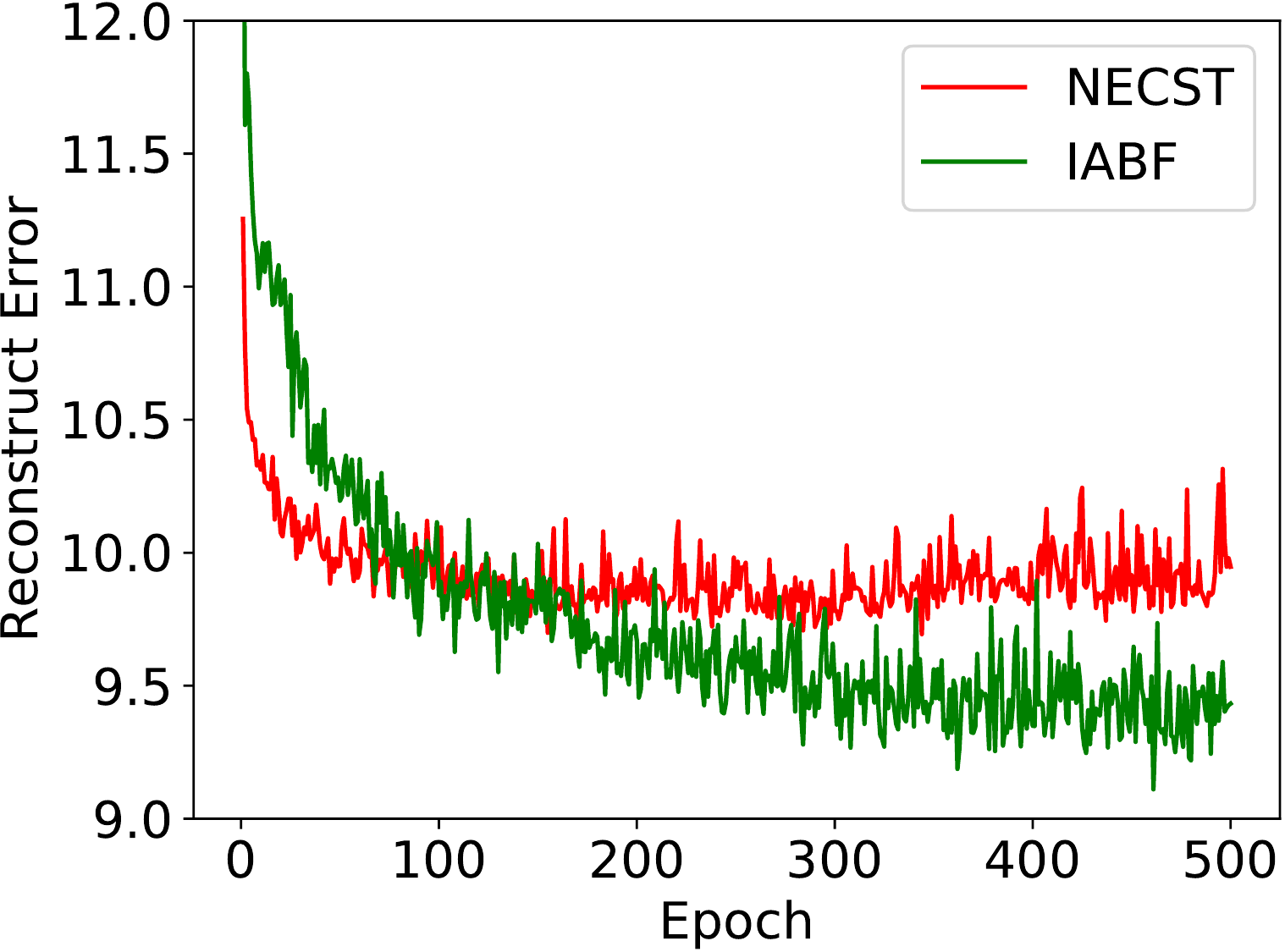}
	\caption{noise level: 0.3}
	\label{fig:noise_3}
\end{subfigure}
\begin{subfigure}{0.49\linewidth}
	\centering
	\includegraphics[width=0.99\columnwidth]{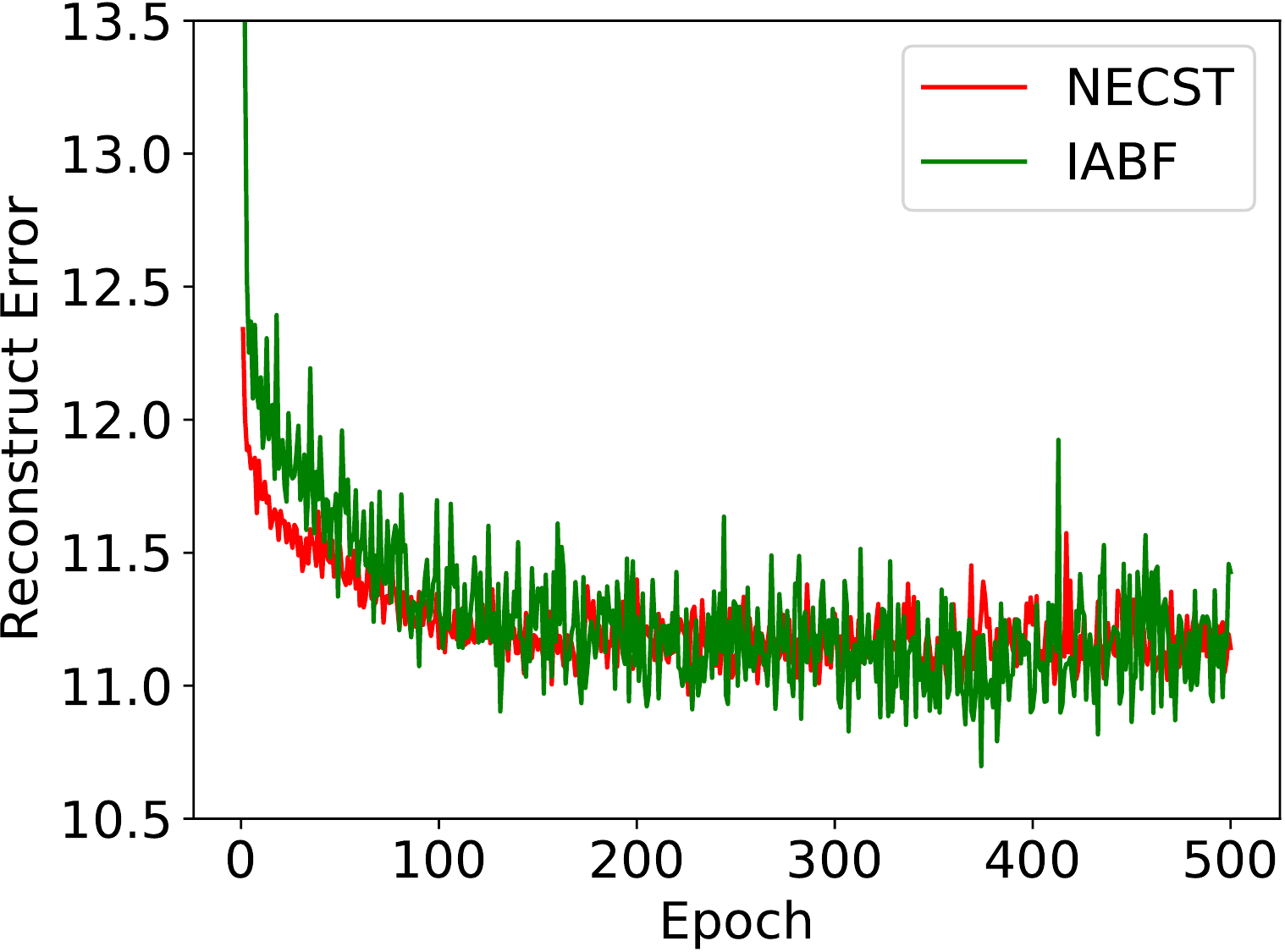}
	\caption{noise level: 0.4}
	\label{fig:noise_4}
\end{subfigure}
\caption{$\mathcal{L}^2$ reconstruction error (per image) of IABF vs. NECST on CIFAR-10 dataset. The error is calculated on validation set during training. Different figure correspond to different noise level. Red: NECST. Green: IABF.}
\label{validation_curve}
\end{figure}

\subsection{Compression and Error Correction}
To validate  our proposed method, we  demonstrate the effectiveness of IABF on compression and error correction and conduct comparison with NECST, which is currently the state-of-the-art joint source-channel coding methods within the finite bit-length setting.

The evaluation setting is directly adapted from NECST~\cite{choi2018necst}. 
With the number of the bits $m$ fixed, we report the corresponding distortion levels(reconstruction errors) on test sets with respect to various noise levels $\epsilon$ in Table.~\ref{compression_results}. The test result is reported according to the model with the lowest distortion on the validation set.
And as observed, IABF can stably outperform the NECST on all binary and RGB datasets within all noise levels from $\epsilon = 0.1$ to $\epsilon = 0.4$, which demonstrates the stability of IABF on dealing with different data complexity and different channel noise level. To verify the effectiveness of different components, we conduct ablation study on MNIST dataset. ABF stands for simply applying Adversarial Bit Flip without mutual information maximization term. It is shown that the ABF itself can already stably outperform NECST, and the with the information maximization term IABF can further decrease the distortion. 

Figure.~\ref{validation_curve} shows the changes of validation reconstruction error with respect to training timesteps for IABF and NECST. We find that the overall training procedure of IABF is more stable than NECST. It is worth noting that at the very beginning of training, the reconstruct error of IABF is higher than NECST. This is due to the fact that at the early stage the encoder is not well trained and hence there is not enough information encoded into the codewords. IABF can more effectively regularize the decoder; however, at this period, the capacity of the well-regularized decoder is not enough to produce good reconstruction.
As the training goes on, NECST shows obvious instability and overfitting phenomenon while the advantages of IABF will emerge. The information maximization part enables the model to achieve less distortion, and the adversarial bits flip helps stable the training and avoid over-fitting. 

\begin{figure}[!t]
\centering     
\begin{subfigure}{0.45\linewidth}
	\centering
	\includegraphics[width=0.99\columnwidth]{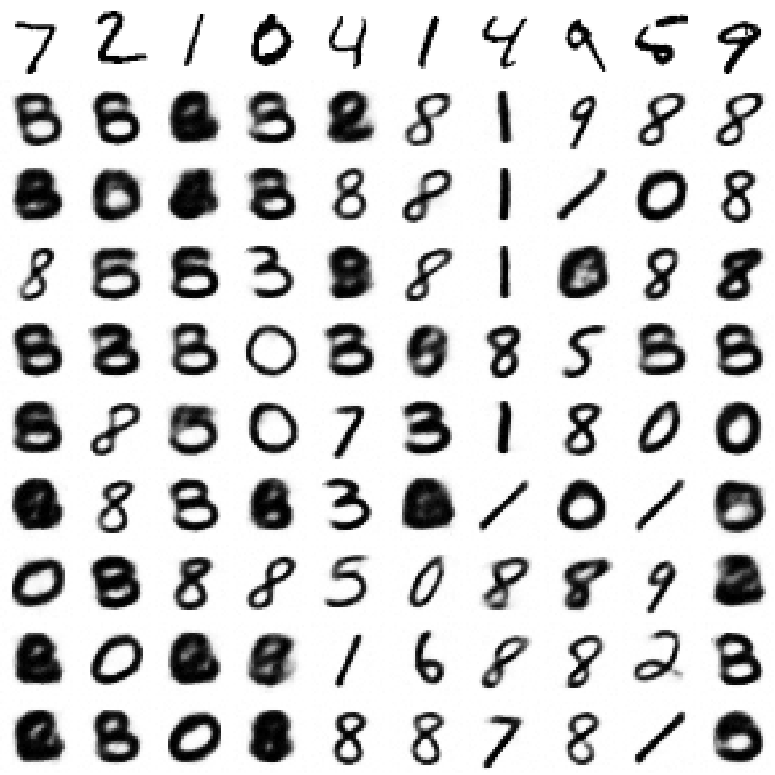}
	\caption{NECST}
	\label{fig:necst_markov_chain_samples}
\end{subfigure}
\hspace{+5pt}
\begin{subfigure}{0.45\linewidth}
	\centering
	\includegraphics[width=0.99\columnwidth]{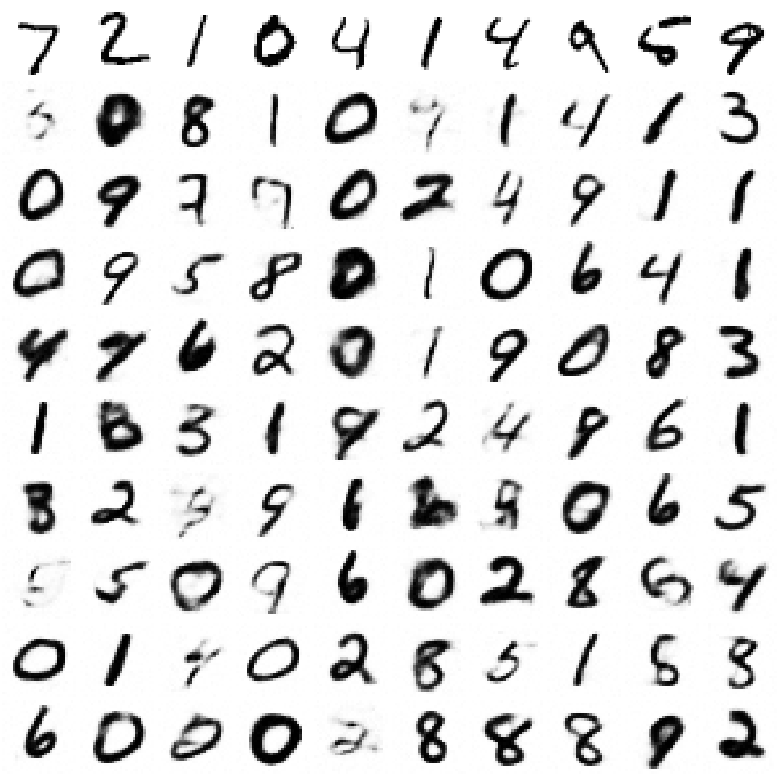}
	\caption{IABF}
	\label{fig:iabf_markov_chain_samples.png}
\end{subfigure}
\caption{Markov chain samples from NECST and IABF. Models are trained on MNIST dataset with 0.1 noise level.}
\label{markov_chain}
\end{figure}

\subsection{Implicit Generative Modeling}
As shown in  \cite{grover2018uncertainty,choi2018necst}, the latent variable model in IABF and NECST  specifies an implicit generative model\cite{mohamed2016learning} under certain conditions. We directly adapt their results here:
starting from any data point $x^{(0)} \sim \mathcal{X}$, we define a Markov chain over $\mathcal{X} \times \mathcal{Y}$ with the following transitions:
\begin{equation}
\label{markov_sampling}
\vy^{(t)} \sim p_{\text{noisy}}\left(\vy | \vx^{(t)} ; \theta, \epsilon\right), \vx^{(t+1)} \sim q_\phi (\vx | \vy^{(t)} )
\end{equation}
We show the samples obtained by running the chain for the model trained with IABF and NECST in Figure.~\ref{markov_chain}. It can be found that the IABF is able to learn a generative model with both better sample quality and diversity than NECST. More specifically, the generative model implied by NECST shows severe model collapse phenomenon due to fact that the information on codewords may be concentrated on several dimensions, which will results in several clusters in the marginal codeword distribution.
Hence during sampling, the sampled data may fall into some clusters. 
While in IABF, both the information maximization term and the adversarial bit flip tend to encourage codewords uniformly distributed, hence there will be fewer clusters and barriers in the latent codeword space which will make the ergodic condition for Markov sampling well satisfied.

\section{Conclusion}
We propose IABF to improve neural joint source-channel coding, where the information theoretic dependency between codewords and data is enhanced without the involvement of parameterized variational distribution and the amortized decoder is also regularized in an adversarial fashion. Experimental results demonstrate that IABF is able to stably improve both the compression and error correction ability of the coding scheme within various kinds of data and noise levels. Active learning may be another option to implement adversarial bits flip, we leave it as future direction.

\section{Acknowledgement}
The corresponding author Shuo Shao thanks the support of National Natural Foundation of China (61872149, 61901261), Shanghai Sailing Program (19YF1424200).
{\small
\bibliographystyle{aaai}
\bibliography{reference}
}
\end{document}